\documentclass[a4paper,11pt]{article}

\title{Are Hitting Formulas Hard for Resolution?}

\author{Tomáš Peitl \\ \texttt{peitl@ac.tuwien.ac.at} \and Stefan Szeider \\ \texttt{sz@ac.tuwien.ac.at}}
\date{}

\usepackage[utf8]{inputenc}
\usepackage{hyperref}
\usepackage{url}
\usepackage{microtype}
\usepackage{enumerate}
\usepackage{amsmath,amsthm}
\usepackage{multirow}
\usepackage{amsfonts}
\usepackage{booktabs}
\usepackage{amssymb}
\usepackage{tikz}
\usepackage{array}
\usepackage{xspace}
\usepackage{flushend}
\usepackage{textcomp}
\usepackage{cite}
\usepackage[left=1.5in,right=1.5in]{geometry}

\usepackage{float}

\usepackage{tikzit,calc}
 
\tikzstyle{var}=[fill=lightgray, draw=black, shape=circle, inner
sep=0pt, minimum size=4pt]
\tikzstyle{cls}=[fill=white, draw=black, shape=rectangle, inner
sep=0pt, minimum size=4pt]

\newcommand{\lv}[1]{}
\newcommand{\sv}[1]{#1}

\newcommand{\n}[1]{\overline{#1}}
\newcommand{\propvar}[2]{\ensuremath{\mathrm{#1}[#2]}}

\newcommand{\prooflen}{s}
\newcommand{\proofseq}{P}
\newcommand{\MUtwo}[1]{\mathcal{F}_{#1}^2}

\newcommand{\Card}[1]{|#1|}

\newcommand{\maps}{\mathrel{\rightarrow}}

\let\phi=\varphi
\let\epsilon=\varepsilon

\newcommand{\CCC}{\mathcal{C}}

\newcommand{\ol}{\overline}

\newcommand{\mtext}[1]{\text{\normalfont\itshape #1}}

\newcommand{\lit}{\mtext{lit}}
\newcommand{\var}{\mtext{var}}

\newcommand{\IUH}{\text{\normalfont IUH}}

\newcommand{\SB}{\{\,}%
\newcommand{\SM}{\mid}
\newcommand{\SE}{\,\}}%

\usepackage{xcolor}
\colorlet{MyBlue}{blue!50!black!100!}
\colorlet{MyRed}{red!50!black!100!}

\newcommand{\modelcount}[1]{{\##1}}
\newcommand{\cset}{\mathcal{C}}
\newcommand{\sir}[1]{{#1^*}}

\newcommand{\Lemma}[1]{Lemma~\ref{lem:#1}}
\newcommand{\Theorem}[1]{Theorem~\ref{thm:#1}}

\newcommand{\orb}{\mathcal{O}}
\newcommand{\symg}[1]{{\mathcal{S}(#1)}}

\newtheorem{definition}{Definition}
\newtheorem{theorem}{Theorem}
\newtheorem{corollary}{Corollary}
\newtheorem{lemma}{Lemma}
\newtheorem{example}{Example}

\definecolor[named]{lipicsYellow}{rgb}{0.99,0.78,0.07}

\begin{document}

\maketitle

\begin{abstract}
  Hitting formulas, introduced by Iwama, are an unusual class of
  propositional CNF formulas. Not only is their satisfiability
  decidable in polynomial time, but even their models can be counted
  in closed form. This stands in stark contrast with other
  polynomial-time decidable classes, which usually have algorithms
  based on backtracking and resolution and for which model counting
  remains hard, like 2-SAT and Horn-SAT. However, those
  resolution-based algorithms usually easily imply an upper bound on
  resolution complexity, which is missing for hitting formulas. Are
  hitting formulas hard for resolution?

  In this paper we take the first steps towards answering this
  question. We show that the resolution complexity of hitting formulas
  is dominated by so-called irreducible hitting formulas, first
  studied by Kullmann and Zhao, that cannot be composed of smaller
  hitting formulas. However, by definition, large irreducible
  unsatisfiable hitting formulas are difficult to construct; it is not
  even known whether infinitely many exist. Building upon our
  theoretical results, we implement an efficient algorithm on top of the
  Nauty software package to enumerate all irreducible unsatisfiable
  hitting formulas with up to 14 clauses. We also determine the exact
  resolution complexity of the generated hitting formulas with up to
  13 clauses by extending a known SAT encoding for our purposes. Our
  experimental results suggest that hitting formulas are indeed hard
  for resolution.
\end{abstract}

\section{Introduction}

Hitting formulas (also known as hitting clause-sets) are propositional formulas in conjunctive normal form (CNF) with the property that each pair of clauses clashes, meaning there is a variable that occurs in one clause and its negation in the other.
Introduced by Iwama \cite{Iwama89} in 1989 and given
the name `hitting' by Kleine B\"{u}ning and Zhao \cite{KleineBuningZhao01a},
they are notorious for Iwama's stunningly
elegant argument (reproduced here as Theorem~\ref{thm:hitting-models}), which shows that their satisfiability can be decided in
polynomial time---in fact, the number of models can be obtained in
\emph{closed form}.  This is a rare encounter in the bay of
tractability: most other polynomial-time decidable subclasses of SAT
(`islands of tractability') have algorithms with more typical
ingredients like unit propagation, (bounded-depth) backtracking, and
resolution~\cite{FrancoMartin21}.  Such algorithms usually do not
scale to the case of model counting, which remains \#P-hard for 2-SAT
and Horn-SAT, two of the most prominent tractable cases of
SAT~\cite{Roth96}. Due to their remarkable properties, hitting formulas
have been the subject of many theoretical investigations
\cite{GalesiKullmann04,KleineBuningZhao01a,KleineBuningZhao03,Kullmann03,Kullmann10,Kullmann11,KullmannZhao16,OrdyniakPaulusmaSzeider13,Szorenyi08,Zhao06}
and served as a building block of larger classes of tractable
formulas via backdoors \cite{NishimuraRagdeSzeider07} and
treewidth~\cite{GanianSzeider21}.

\begin{figure}
\[ \{x, \n{y}\}, \{y, \n{z}\}, \{z, \n{x}\}, \{x, y, z\}, \{\n{x}, \n{y}, \n{z}\} \]
\caption{An example hitting formula written as a set of clauses, each itself a set of literals.}
\label{fig:hitting-formula}
\end{figure}

But one fundamental question remains unsettled: while polynomial-time
resolution-like algorithms usually provide a corresponding polynomial
upper bound on the \emph{resolution complexity} (length of a shortest resolution refutation) of formulas they
target, it is far from clear how one could get such a bound from the counting argument
for hitting formulas.
Resolution complexity has established itself as an important measure due to its relation to the running time of conflict-driven clause-learning (CDCL~\cite{MarquessilvaSakallah96}) SAT solvers---resolution complexity gives a lower bound for CDCL running time and, under some assumptions also an upper bound (although these assumptions include non-determinism)~\cite{BeameKautzSabharwal04,AtseriasFichteThurley11,PipatsrisawatDarwiche09}.
Even though the resolution complexity of hitting formulas might not have immediate practical consequences, it is still a natural question to ask for a famous class of formulas and a fundamental complexity measure.

So, are hitting formulas hard for resolution?  On the one hand, they
are easy even for model counting; on the other hand, the reason why
they are easy to solve is a global counting argument, the kind which
is notoriously hard for resolution (as witnessed, for instance, by the
Pigeonhole Principle~\cite{Haken85} or Tseitin Formulas \cite{Urquhart87}).  With
this paper, we attempt to shed some light on this question.

One reason why we know so little about the resolution complexity of hitting formulas is that we have no good methods of constructing large hard instances.
We can construct large hitting formulas by carefully gluing together smaller hitting formulas, but that only produces \emph{reducible} formulas, a notion recently proposed by Kullmann and Zhao~\cite{KullmannZhao16}.
In contrast, we observe (Corollary~\ref{cor:riuh-hard}) that any sequence of hitting formulas must implicitly contain arbitrarily large \emph{irreducible} formulas  to have a chance at being hard for resolution.
However, large irreducible formulas are non-trivial to construct because they
cannot be assembled from smaller ones by definition, at least not by known constructions.

That leads us to another straightforward yet non-trivial
question: do there even \emph{exist} infinitely many irreducible
unsatisfiable hitting formulas?  We hint that this is indeed the case
by enumerating all irreducible unsatisfiable hitting formulas on up to 14 clauses
(and some larger) modulo isomorphisms and observing that their numbers grow fast.
Kullmann and Zhao proved~\cite[Lemma 39]{KullmannZhao16}, perhaps suprisingly,
that all irreducible unsatisfiable hitting formulas other than
$x \wedge \neg x$ must be \emph{regular}---meaning each literal must
occur at least twice (also known as \emph{nonsingular}, though we prefer the more positive term).
Another surprise we discovered is that there is no irreducible unsatisfiable hitting formula with 4 or 6 clauses
(and none with 2 or 3, but that is an artifact of regularity; see
Table~\ref{table:IUH-hardness}).

Now that we have the formulas, we might as well compute their shortest
refutations.  Armed with recent developments in both the theory of hitting
formulas~\cite{KullmannZhao16} as well as in practical computation of
shortest resolution refutations~\cite{PeitlSzeider21}, we can compute
shortest refutations of all irreducible hitting formulas on up to $13$
clauses (and some larger) and all regular hitting formulas on up to
$11$ clauses.  The picture that this unveils indicates that hitting
formulas are perhaps not as hard as the hardest formulas in general
(by comparing with the \emph{resolution hardness
  numbers}~\cite{PeitlSzeider21}), but not significantly easier
either.

We need some theoretical advances to narrow our search space and
compute shortest refutations faster.  Following the lead of Kullmann
and Zhao~\cite{KullmannZhao13,KullmannZhao16}, we focus on the
aforementioned concept of \emph{irreducibility}, and we observe that a
new property we call \emph{strong irreducibility} is also essential.
We prove irreducibility and strong irreducibility coincide for
hitting formulas, allowing us to take advantage of the stronger
variant while only checking the weaker one.

Our contributions are thus twofold.  First, we contribute theoretical
knowledge about irreducibility and regularity in (not only) hitting
formulas, with a focus on connections to resolution complexity.  Second, we extensively evaluate the
resolution complexity of hitting formulas using the new theoretical
results. In the experimental part, we rely on two key tools: the
ability to generate formulas modulo isomorphisms, provided by a
customized version of Nauty~\cite{Nauty}, and an efficient way to
compute shortest refutations of generated formulas.  For the latter, we use
a recently proposed SAT encoding~\cite{PeitlSzeider21}, but we enhance
it with further symmetry breaking and constraints that exploit
irreducibility.

Our findings indicate that small hitting formulas are almost as hard
for resolution as the hardest formulas with the same number of
clauses, suggesting that hitting formulas in general might not admit
polynomial-size resolution refutations.


\section{Preliminaries}
\label{sec:prelims}

We assume familiarity with standard notions of graph theory, including
those of \emph{(un)directed graphs}, \emph{acyclicity}, and \emph{in-}
and \emph{out-degree} of a vertex---we refer to a standard graph
theory handbook~\cite{BondyMurty08}. All graphs considered, directed
or undirected, do not contain any self-loops or parallel edges.

\subsection{Formulas}
We consider propositional formulas in conjunctive normal form (CNF)
represented as sets of clauses. We assume an infinite set $\mathcal{V}$ of
(propositional) \emph{variables}.  A \emph{literal} $\ell$ is a
variable $x$ or a negated variable $\neg x$; we write $\lit:=\SB x,
\neg x\SM x\in \mathcal{V}\SE$.  For a literal $\ell$ we write $\ol{\ell}:=\neg
x$ if $\ell=x$, and $\ol{\ell}:=x$ if $\ell=\neg x$.  For a set $C$ of
literals, we define $\ol{C}:=\SB\ol{\ell}\SM \ell\in C\SE$; we say $C$ is \emph{tautological} if $C\cap\ol{C}\neq
\emptyset$. A finite non-tautological set of literals is a
\emph{clause}; a finite set of clauses is a (CNF) \emph{formula}.
The empty clause is denoted by $\bot$.
We say two clauses $C$ and $D$ \emph{clash} if $C \cap \n{D} \neq \emptyset$; the literals in $C \cap \n{D}$ and $\n{C} \cap D$ are the \emph{clashing literals}, and their variables are the \emph{clashing variables}. 
For a clause $C$, we put $\var(C) = \SB \var(\ell) \SM \ell \in C \SE$, and for a formula $F$, $\var(F) = \bigcup_{C \in F} \var(C)$.
Similarly, we put $\lit(F):=\var(F)\cup \ol{\var(F)}$.

An \emph{(partial) assignment} is a mapping $\tau:V \subseteq \var(F)\maps \{0,1\}$, extended to literals as $\tau(\n{x}) = 1-\tau(x)$.
An assignment is called \emph{total} if its domain is $\var(F)$.
We identify an assignment with the set of literals $\tau^{-1}(1) \cup \n{\tau^{-1}(0)}$, i.e., with the set of all literals set to $1$ by $\tau$.
For a clause $C$ and a set of literals $\tau$ (which could be an assignment) we define the \emph{restriction} of $C$ by $\tau$, written $C[\tau]$, as $\top$ if $C \cap \tau \neq \emptyset$ (in which case we say $\tau$ \emph{satisfies} $C$), and $C \setminus \n{\tau}$ otherwise.
If $C[\tau] = \bot$ for a clause $C$ and total assignment $\tau$, we say $C$ \emph{covers} $\tau$.
The \emph{restriction} of a formula $F$ by an assignment $\tau$, written $F[\tau]$, is defined as $\{C[\tau] \SM C \in F\} \setminus \{\top\}$.
A formula $F$ is \emph{satisfiable} if there is a \emph{satisfying assignment} (also \emph{model}), i.e., a total assignment $\tau$ with $F[\tau] = \emptyset$; otherwise it is \emph{unsatisfiable}.
For two formulas $F$ and $G$, we say $F$ \emph{entails} $G$, denoted by $F \models G$, if every model of $F$ is a model of~$G$.
We say $F$ and $G$ are \emph{(logically) equivalent}, denoted $F \equiv G$, if $F \models G$ and $G \models F$.
We write $\modelcount{F}$ for the number of models of $F$.

\subsection{Resolution Refutations}
If $C_1\cap\ol{C_2}=\{\ell\}$ for clauses $C_1,C_2$ and a literal
$\ell$, then the \emph{resolution rule} allows the derivation of the
clause $D=(C_1\cup C_2) \setminus \{\ell,\ol{\ell}\}$; $D$ is the
\emph{resolvent} of the \emph{premises} $C_1$ and $C_2$, and we say that $D$ is obtained
by \emph{resolving on~$\ell$}.  For a formula $F$,
a sequence $\proofseq=C_1,\dots,C_\prooflen$ of clauses is a
\emph{resolution derivation of $C_\prooflen$ from $F$} if for each
$i\in \{1,\dots,\prooflen\}$ at least one of the following holds.
\begin{enumerate}
\item  $C_i\in F$ (``$L_i$ is an axiom'');
\item  $C_i$ is the resolvent of $L_j$ and $L_{j'}$ for some $1\leq
  j<j'<i$ (``$C_i$ is obtained by resolution'').
\end{enumerate}
We write $\Card{\proofseq}:=\prooflen$ and call $\prooflen$ the \emph{length} of $\proofseq$.  If
$L_\prooflen$ is the empty clause, then $\proofseq$ is a \emph{resolution refutation} of~$F$.
A clause $C_i$ in a resolution
derivation may have different possible `histories.' $C_i$ may
be the resolvent of more than one pair of clauses preceding $C_i$, or
$C_i$ may be both an axiom and obtained from preceding clauses by
resolution. In the sequel, however, we assume that an arbitrary
but fixed single history is associated with each considered resolution
derivation.
Thus, with a refutation $\proofseq$ we can associate the directed acyclic graph (\emph{refutation DAG}) $G(\proofseq)$ whose vertices are the clause of $\proofseq$, and which has an arc from $C_i$ to $C_j$ if there is $C_k$, $i, k < j$, such that $C_j$ is the resolvent of $C_i$ and $C_k$.
In any refutation DAG $G(\proofseq)$, the in-degree of each axiom vertex is~0, while each resolvent has in-degree 2.

It is well known that resolution is a complete proof system for
unsatisfiable formulas: a formula is unsatisfiable if and
only if there exists a resolution refutation of it~\cite{DavisPutnam60}.  The
\emph{resolution complexity} or \emph{resolution hardness} $h(F)$ of
an unsatisfiable formula $F$ is the length of a shortest resolution
refutation of~$F$.
For a nonempty set $\CCC$ of formulas, we define
$h(\CCC) = \sup_{F \in \CCC} h(F)$.

\subsection{Unsatisfiability Minimal and Saturated}

A formula $F$ is \emph{minimally unsatisfiable} if it is unsatisfiable and removing any clause $C \in F$ leaves $F \setminus \{C\}$ satisfiable.
A formula $F$ is  \emph{saturated minimally unsatisfiable} if it is unsatisfiable and adding any literal (including on a fresh variable) to any clause makes it satisfiable.
The central notion of the paper is, of course, that of a \emph{hitting formula}.

\begin{definition}
	A formula $F$ is \emph{hitting} if for every $C, D \in F$, $C \neq D$, $C$ and $D$ clash.
\end{definition}

Hitting formulas are notable for the following useful property.

\begin{theorem}[Iwama \cite{Iwama89}]
	\label{thm:hitting-models}
	The number of satisfying assignments of a hitting formula $F = \{C_1, \dots, C_m\}$ with $n$ variables is given by $2^n(1 - \sum_{i=1}^m 2^{-|C_i|})$.
\end{theorem}
\begin{proof}
	Because each pair of clauses of $F$ clashes, every assignment can only be covered by at most one clause.
	Since for each clause we can explicitly calculate the number of assignments it covers, we get the overall number of covered assignments---models are simply assignments that are not covered.
\end{proof}

\begin{corollary}
  A hitting formula $F$ is unsatisfiable if and only if
  $\sum_{C \in F} 2^{-|C|} = 1$.
\end{corollary}

\begin{corollary}
  All unsatisfiable hitting formulas are saturated minimally
  unsatisfiable.
\end{corollary}
\begin{proof}
	Consider what happens to the model-counting sum after adding a literal or deleting a clause (both of which are operations that trivially preserve hittingness).
\end{proof}

Two important notions in the context of minimally unsatisfiable
formulas are \emph{singularity} and \emph{deficiency}. \lv{

  }A literal $\ell \in \lit(F)$ is \emph{singular} if it is only contained in one clause.
A variable $v \in \var(F)$ is singular if $v$ or $\n{v}$ is a singular literal.
A formula is singular if it has a singular variable, and is \emph{regular} otherwise.
Singular formulas can be reduced by \emph{DP-reduction}---the elimination of a variable $v$ by taking all possible resolvents on $v$ and then removing all clauses that contained $v$ or $\n{v}$.
While DP-reduction can be applied to any variable of any formula preserving satisfiability, \emph{singular} DP-reduction---when the eliminated variable $v$ is singular---always decreases formula size, and thus can be applied exhaustively until a regular formula is obtained in overall polynomial time.
Thus arises the importance of regular formulas---they are a kind of `core' left over once singular variables are eliminated.

The other important notion in the context of minimally unsatisfiable formulas is that of the \emph{deficiency}, defined as the difference between the number of clauses and the number of variables of a formula, and known to be positive for minimally unsatisfiable formulas~\cite{AharoniLinial86}.
It is known that minimally unsatisfiable formulas of deficiency $1$ with more than $1$ clause are always singular, and, up to isomorphism, there is only one regular minimally unsatisfiable formula of deficiency $2$ for each number of clauses $m \geq 4$~\cite{KleineBuning00} (otherwise there are none); we will refer to it as $\MUtwo{m}$.
These formulas are actually saturated minimally unsatisfiable, and for $m=4,5$ also hitting.
Kullmann and Zhao refer to $\MUtwo{4}$ and $\MUtwo{5}$ as $\mathbf{\mathcal{F}}_2$ and $\mathbf{\mathcal{F}}_3$ respectively~\cite{KullmannZhao16}.

It is known that singular DP-reduction of minimally unsatisfiable formulas preserves both minimal unsatisfiability as well as deficiency, and it also preserves saturated minimal unsatisfiability~\cite[Corollaries 10 and 13]{KullmannZhao13}.
While singular DP-reduction is not necessarily confluent in minimally unsatisfiable formulas (different orders of elimination can result in different formulas), it is confluent in saturated minimally unsatisfiable formulas~\cite[Thm 23]{KullmannZhao13}.

We will use the following acronyms to describe various classes of formulas: UH (unsatisfiable hitting), RUH (regular UH), IUH (irreducible UH, necessarily regular), SMU (saturated minimally unsatisfiable), RSMU (regular SMU), SSMU (singular SMU), RISMU (irreducible RSMU), SISMU (irreducible SSMU).
For each of these classes we may append one or two integral parameters in parentheses: for example $\IUH(n, m)$ are IUHs with exactly $n$ variables and $m$ clauses, and $\IUH(m)$ are IUHs with exactly $m$ clauses (without the restriction on the number of variables).

\subsection{Isomorphisms and Symmetries}

\newcommand{\Aut}{\mathsf{Aut}}
\newcommand{\Orb}{\mathsf{Orb}}
\newcommand{\Stb}{\mathsf{Stb}}

An \emph{isomorphism} between two formulas $F$ and $G$ is a
bijection $\phi: \lit(F) \rightarrow \lit(G)$ such that for each
literal $\ell \in \lit(F)$ we have $\ol{\phi(\ell)}=\phi(\ol{\ell})$
and for each $C\subseteq \lit(F)$ we have $C\in F$ if and only if
$\phi(C) := \SB \phi(\ell) \SM \ell \in C\SE\in G$.
If there is an isomorphism between $F$ and $G$, we say $F$ and $G$ are \emph{isomorphic}.
For instance the formulas $F =\{\{x,y\}$,
$\{\ol{x},y\}$, $\{\ol{y}\}\}$, and $G=\{\{\ol{z},w\}$,
$\{z,w\}$, $\{\ol{w}\}\}$ are isomorphic.
Obviously, two isomorphic formulas have the same properties concerning
satisfiability, minimal unsatisfiability, and resolution complexity.

An \emph{automorphism} (or symmetry)
of a CNF formula $F$ is an isomorphism to itself.
$\Aut(F)$ denotes the set of automorphisms of $F$.
For an automorphism $\phi\in \Aut(F)$ we denote by $\phi^*$ the
bijection $\var(F)\rightarrow \var(F)$ defined by
$\phi^*(x)=\var(\phi(x))$; $\phi^*$ is the \emph{variable action} of
$\phi$ on~$x$.
Let $\Aut^*(F)$ denote the set of all variable actions of
$\Aut(F)$.

The \emph{orbit} of a literal $\ell$ in $F$ is the set
$\Orb_F(\ell)=\SB \phi(\ell) \SM \phi\in \Aut(F)\SE$.  The
\emph{variable orbit} of a variable $x$ in $F$ is the set
$\Orb_F^*(x)=\SB \phi^*(x) \SM \phi\in \Aut^*(F)\SE$.
\lv{A formula is called \emph{variable-transitive} if the variable action has a single orbit.}


For sets $S, S', T$, we write  $T = S \uplus S'$ if $T = S \cup S'$ and $S \cap S' = \emptyset$.
A \emph{2-graph} is an undirected graph $G=(V,E)$ together with a
partition of its vertex set into two disjoint subsets $V_1 \uplus V_2 = V$.  Two
2-graphs $G=(V_1 \uplus V_2,E)$ and $G'=(V_1' \uplus V_2',E')$ are
\emph{isomorphic} if there exists a bijection
$\phi: V_1 \uplus V_2 \rightarrow V_1' \uplus V_2'$ such that
$v\in V_i$ if and only if $\phi(v)\in V_i'$, $i=1,2$, and
$\{u,v\} \in E$ if and only if $\{\phi(u),\phi(v\}\} \in E'$.

The \emph{clause-literal graph} of a formula $F$ is the 2-graph $G(F)=(V_1 \uplus V_2,E)$ with $V_1=\lit(F)$, $V_2= F$, and 
$E=\SB \{x,\ol{x}\} \SM x\in \var(F)\SE \cup \SB \{C,\ell\} \SM C\in
F, \ell \in C \SE$.
We refer to the edges $\SB x, \n{x} \SE$ as \emph{variable} edges.
%
%
%
It is easy to verify that any two formulas are isomorphic if and only
if their clause-literal graphs are isomorphic (as 2-graphs).

\section{Irreducibility}
\sv{\enlargethispage*{5mm}}


The central notion of this section is that of a formula \emph{factor} (called `clause-factor', and later `clause-irreducible' by Kullmann and Zhao~\cite{KullmannZhao16}).

\begin{definition}[Kullmann and Zhao \cite{KullmannZhao16}]
	A \emph{factor} of a formula $G$ is a subset $F \subseteq G$ logically equivalent to a single clause $C$.
	The clause $C$ is called the \emph{basis} of the factor $F$ and is uniquely determined.
\end{definition}

\begin{lemma}
	Let $F \subseteq G$.
	Then $F$ is a factor if and only if there is a clause $C$ such that $C \subseteq D$ for every $D \in F$ and $\SB D \setminus C \SM D \in F \SE$ is unsatisfiable. 
\end{lemma}
\begin{proof}
	If $F$ is a factor, then its basis satisfies the conditions on $C$.
	Conversely, if there is such a $C$, then $F$ must be equivalent to it.
\end{proof}

Non-trivial formula factors, like non-trivial factors for integers and other kinds of algebraic objects, can be used to decompose or reduce formulas.
That leads to the central notion of an \emph{irreducible} formula---one which cannot further be reduced by collapsing factors to their bases.

\begin{definition}[Kullmann and Zhao \cite{KullmannZhao16}]
	A formula $G$ is called \emph{irreducible} if for each factor $F \subseteq G$ either $|F| = 1$ or $F = G$.
	Otherwise, it is reducible.
\end{definition}

There is an important connection between factors and resolution hardness.
In a nutshell, in presence of non-trivial factors, one can always pursue the strategy of reducing any remaining non-trivial factors to their bases, gradually shrinking the formula.
These reductions can be performed in resolution, giving a way of constructing a canonical `decomposition' refutation.
An upper bound on resolution complexity follows.

\begin{lemma}
	\label{lem:hardness-bound}
	Let $G$ be an unsatisfiable formula with $m$ clauses, $F \subseteq G$ a factor of size $k$, $C$ the basis of $F$.
	Then $G$ has a proof of length at most $h(F[\n{C}]) + h( \{C\} \cup G \setminus F) - 1$.
\end{lemma}
\begin{proof}
	Because $F$ is a factor, $F[\n{C}]$ is unsatisfiable and can be refuted in $h(F[\n{C}])$ resolution steps.
	By re-inserting $C$, we can derive $C$ from $F$ in $h(F[\n{C}])$ resolution steps.
	Because $F \equiv C$, $G \setminus F \cup \{C\}$ is still unsatisfiable, and hence can be refuted in $h( \{C\} \cup G \setminus F)$ steps.
	Combining the two refutations minus $1$ for the shared clause $C$ we get the result.
\end{proof}

In the previous proof, we do not actually need that $F \equiv C$.
All we need is that $F \models C$ and $G \setminus F \cup C$ is unsatisfiable.
In other words, that $C$ is a kind of `interpolant' of $F$ and $G \setminus F$.
Such generalised factors will turn out quite useful in the context of resolution.

\begin{definition}
	A clause $C$ is a \emph{clause interpolant} of $(F, G)$ if $F \models C$ and $C \models \neg G$.
	We say that $F \subseteq H$ is a \emph{pseudo-factor} of $H$ if $(F, H \setminus F)$ has a clause interpolant.
	A formula $H$ with $m$ clauses is \emph{strongly irreducible} if it has no pseudo-factors of size other than $1$ or $m$, otherwise it is \emph{weakly reducible}.
\end{definition}

\begin{corollary}
	\label{cor:hardness-bound}
	Let $G$ be an unsatisfiable formula with $m$ clauses, $F \subseteq G$ a pseudo-factor of size $k$, $C$ a clause interpolant of $(F, G \setminus F)$.
	Then $G$ has a refutation of length at most $h(F[\n{C}]) + h( \{C\} \cup G \setminus F) - 1$.
\end{corollary}

It is easy to see that strong irreducibility is indeed a strengthening of irreducibility (\Lemma{strongly-is-irreducible}), but the converse is not true even for MU formulas (Example~\ref{ex:irreducible-is-not-strongly}).

\begin{lemma}
	\label{lem:strongly-is-irreducible}
	Every unsatisfiable strongly irreducible formula is irreducible.
\end{lemma}
\begin{proof}
	A non-trivial factor is by definition also a non-trivial pseudo-factor.
\end{proof}

\begin{example}
	\label{ex:irreducible-is-not-strongly}
	The formula
	$ \{\{x, y\}, \{\n{x}\}, \{\n{y}\}\} $
	is minimally unsatisfiable and irreducible (no pair of clauses is equivalent to a single clause).
	But it is not strongly irreducible: the first two clauses form a pseudo-factor with the clause interpolant $\{y\}$.
\end{example}

The formula in the previous example is minimally unsatisfiable but not saturated.
As it turns out, it must be; for saturated minimally unsatisfiable formulas (which includes hitting formulas), irreducibility implies strong irreducibility.

\begin{theorem}
	\label{thm:strongly-irreducible}
	Any pseudo-factor of a saturated minimally unsatisfiable formula is also a factor.
	Consequently, an SMU formula is strongly irreducible if and only if it is irreducible.
\end{theorem}
\begin{proof}
	For contradiction, let $F \subset G$ be a non-trivial pseudo-factor of the pseudo-reducible SMU formula $G$, let $C$ be the clause interpolant.
	Let $F' = \SB D \cup C \SM D \in F \SE$.
	Then $F' \equiv F \lor C$, and hence $F' \models C$ and so $F' \cup G \setminus F$ is unsatisfiable.
	Because $G$ is saturated, $F' = F$.
	Clearly $C \models F' = F$, and so $F$ is in fact a factor and $G$ is reducible.
\end{proof}

\Theorem{strongly-irreducible} is an important background link that allows us to benefit from strong irreducibility while only having to check irreducibility, which is simpler (cf.\ Section~\ref{sec:generate}, where we describe how we generate irreducible hitting formulas, and Subsection~\ref{subsec:irreducibility-constraint}, where we use strong irreducibility in computing shortest resolution refutations).

Although saturated minimal unsatisfiability is sufficient to elevate irreducibility onto strong irreducibility, it is not necessary.
\begin{example}
\lv{	The formula
	\[ \{\{x_1, \n{x_2}\}, \{x_2, \n{x_3}\}, \{x_3, \n{x_4}\}, \{x_4, \n{x_1}\}, \{x_1, x_2, x_3\}, \{\n{x_1}, \n{x_2}, \n{x_3}, \n{x_4}\}\} \]
}
\sv{\sloppypar	The minimally unsatisfiable formula
	$\{\{x_1, \n{x_2}\}$, $\{x_2, \n{x_3}\}$, $\{x_3, \n{x_4}\}$, $\{x_4, \n{x_1}\}$, $\{x_1, x_2, x_3\}$, $\{\n{x_1}, \n{x_2}, \n{x_3}, \n{x_4}\}\}$ 
}
	is not saturated (it is obtained by deleting the literal $x_4$ from the 5th clause of the unique deficiency-2 RSMU formula $\MUtwo{6}$), yet it is strongly irreducible.
	Admittedly, why it is strongly irreducible is not obvious: we verified this by computer, trying every possible split and every possible clause interpolant.
\end{example}

Finally, we arrive at the statement that we will actually use in our encoding to prune the search spaces by asserting certain kinds of refutations do not exist for strongly irreducible minimally unsatisfiable formulas.

\begin{lemma}
	\label{lem:axiom-reuse}
	Let $F$ be a strongly irreducible minimally unsatisfiable formula with more than $2$ clauses, let $P$ be a resolution refutation of $F$.
	Let $C, D \in F$ be two axioms that are resolved together in $P$.
	Then at least one of $C, D$ is used at least once more in $P$.
\end{lemma}
\begin{proof}
	$\{C , D\} \subsetneq F$ is not unsatisfiable, because $F$ is minimally unsatisfiable.
	Hence $P$ continues beyond the resolution of $C$ and $D$.
	If neither $C$ nor $D$ is used another time, then the resolvent of $C$ and $D$ is a clause interpolant for $(\{C, D\}, F \setminus \{C, D\})$, contradicting $F$'s strong irreducibility.
\end{proof}

Kleine Büning and Zhao~\cite{KleineBuningZhao02b} studied \emph{read-once} refutations for minimally unsatisfiable formulas---ones in which every clause is only used once.
\Lemma{axiom-reuse} has implications for the existence of read-once refutations for strongly irreducible formulas.

\begin{corollary}
	\label{cor:not-read-once}
	Strongly irreducible minimally unsatisfiable formulas with more than $2$ clauses do not have read-once refutations.
\end{corollary}

\tikzstyle{axiom}=[draw=lipicsYellow, thick, solid, rounded corners=4]

\begin{example}
	\label{ex:read-twice}
	Corollary~\ref{cor:not-read-once} is in a sense tight, because there are strongly irreducible formulas with refutations where only a single axiom is read twice and everything else only once.
	An example is the formula
	\[\MUtwo{5}=\{\{x_1, \n{x_2}\}, \{x_2, \n{x_3}\}, \{x_3, \n{x_1}\}, \{x_1, x_2, x_3\}, \{\n{x_1}, \n{x_2}, \n{x_3}\}\},\]
	known to have hardness $10$ by~\cite{PeitlSzeider21}, with a shortest refutation shown in Figure~\ref{fig:read-twice}.

	\begin{figure}
		\sv{\begin{tikzpicture}\sv{[yscale=0.8]}}
		\lv{\begin{tikzpicture}}
			\node[style=axiom] (AP)  at (  -2,  0  ) {$\{x_1, x_2, x_3\}$}  ;
			\node[style=axiom] (A1)  at (  -4,  0  ) {$\{x_1, \n{x_2}\}$}  ;
			\node[style=axiom] (A2)  at (   0,  0  ) {$\{x_2, \n{x_3}\}$}  ;
			\node[style=axiom] (A3)  at (   4,  0  ) {$\{x_3, \n{x_1}\}$}  ;
			\node[style=axiom] (AN)  at (   2,  0  ) {$\{\n{x_1}, \n{x_2}, \n{x_3}\}$}  ;

			\node[] (L1) at (  -1, -1) {$\{x_1, x_2\}$}  ;
			\draw[->] (A2) -- (L1);
			\draw[->] (AP) -- (L1);

			\node[] (L2) at (  -3, -1.5) {$\{x_1\}$}  ;
			\draw[->] (L1) -- (L2);
			\draw[->] (A1) -- (L2);

			\node[] (L3) at (   1, -1) {$\{\n{x_1}, \n{x_3}\}$}  ;
			\draw[->] (A2) -- (L3);
			\draw[->] (AN) -- (L3);

			\node[] (L4) at (   3, -1.5) {$\{\n{x_1}\}$}  ;
			\draw[->] (A3) -- (L4);
			\draw[->] (L3) -- (L4);

			\node[] (L5) at (   0, -2.2) {$\bot$}  ;
			\draw[->] (L2) -- (L5);
			\draw[->] (L4) -- (L5);
		\end{tikzpicture}
		\caption{A shortest resoluton refutation of the
			formula $\MUtwo{5}$ from
		Example~\ref{ex:read-twice} (axioms highlighted).}
				  \label{fig:read-twice}
	\end{figure}
\end{example}

One could ask whether the decomposition refutation from Lemma~\ref{lem:hardness-bound} is optimal.
This is not the case, as witnessed by the following example.

\newcommand{\sep}{\hspace{0.4em}}
\newcommand{\vsep}{\\}
\begin{example}
  \label{ex:decomposition-suboptimal}
  \sv{Let $F = \{
			\{x, y, z\}$,
			$\{\n{x}, \n{y}, \n{z}\}$,
			$\{\n{x}, y, e\}$,
			$\{x, \n{y}, \n{e}\}$,
			$\{x, \n{z}, e\}$,
			$\{\n{x}, z, \n{e}\}$,
			$\{y, \n{z}, \n{e}\}$,
			$\{\n{y}, z, e\}
		\}$. }%
        \lv{
	Let
	\[ F = \{
			\{x, y, z\},
			\{\n{x}, \n{y}, \n{z}\},
			\{\n{x}, y, e\},
			\{x, \n{y}, \n{e}\},
			\{x, \n{z}, e\},
			\{\n{x}, z, \n{e}\},
			\{y, \n{z}, \n{e}\},
			\{\n{y}, z, e\}
		\}
	\]
	written in `incidence matrix' form as (rows are variables, columns are clauses, `+' means positive occurrence, `--' means negative occurrence)
	\begin{center}
		\begin{tabular}{|@{\sep}c@{\sep}c@{\sep}c@{\sep}c@{\sep}c@{\sep}c@{\sep}c@{\sep}c@{\sep}|}
		\hline
		+  & -- & -- & +  & +  & -- &    &    \vsep
		+  & -- & +  & -- &    &    & +  & -- \vsep
		+  & -- &    &    & -- & +  & -- & +  \vsep
		   &    & +  & -- & +  & -- & -- & +  \vsep
		\hline
	\end{tabular}
      \end{center}
    }%
$F$ is the formula $F_{4, 8, 52}$ from \cite{PeitlSzeider21} and so is known to have hardness $19$.
	Now, consider $G$ obtained from $F$ by replacing the clause $\{\n{x}, z, \n{e}\}$ with the clauses $\{\n{x}, y, z, \n{e}\}$ and $\{\n{x}, \n{y}, z, \n{e}\}$.
	Clearly, $G$ is reducible: the two new clauses are a factor, whose basis is the replaced clause.
	If the decomposition refutation were optimal, the hardness of $G$ would have been $3 + h(F) - 1 = 21$.
	But $G$ has a refutation of length $20$ (see Figure~\ref{fig:refu20}).
	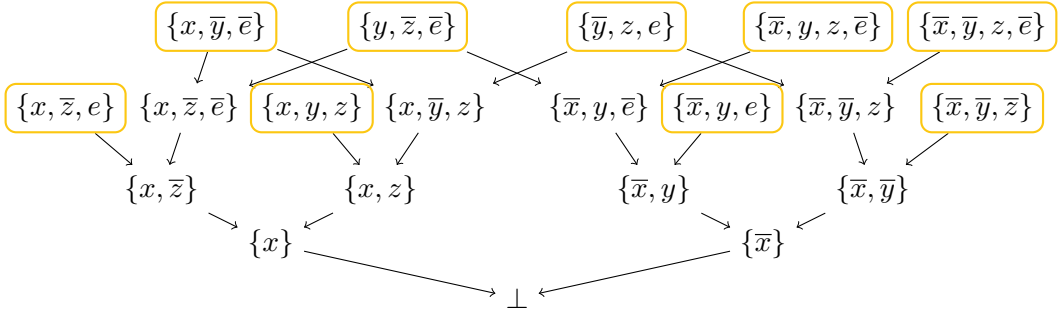
\begin{figure}
		\begin{tikzpicture}[scale=0.36]
			\begin{pgfonlayer}{nodelayer}
				\node [style=axiom] (A1) at (-8.5, 2) {$\{   x ,    y ,    z        \}$};
				\node [style=axiom] (A2) at (6.5, 2) {$\{\overline{x},    y ,           e \}$};
				\node [style=axiom] (A7) at (-17.5, 2) {$\{   x ,        \overline{z},    e \}$};
				\node [style=axiom] (A6) at (-12, 5) {$\{   x , \overline{y},        \overline{e}\}$};
				\node [style=axiom] (A5) at (3, 5) {$\{       \overline{y},    z ,    e \}$};
				\node [style=axiom] (A3) at (-5, 5) {$\{          y , \overline{z}, \overline{e}\}$};
				\node [style=axiom] (A8) at (10, 5) {$\{\overline{x},    y ,    z , \overline{e}\}$};
				\node [style=axiom] (A9) at (16, 5) {$\{\overline{x}, \overline{y},    z , \overline{e}\}$};
				\node [style=axiom] (A4) at (16, 2) {$\{\overline{x}, \overline{y}, \overline{z}       \}$};
				\node [] (L10) at (-4, 2) {$\{   x , \overline{y},    z        \}$};
				\node [] (L11) at (-6, -1) {$\{   x ,           z        \}$};
				\node [] (L12) at (-13, 2) {$\{   x ,        \overline{z}, \overline{e}\}$};
				\node [] (L13) at (-14, -1) {$\{   x ,        \overline{z}       \}$};
				\node [] (L14) at (-10, -3) {$\{   x                      \}$};
				\node [] (L15) at (2, 2) {$\{\overline{x},    y ,        \overline{e}\}$};
				\node [] (L16) at (4, -1) {$\{\overline{x},    y               \}$};
				\node [] (L17) at (11, 2) {$\{\overline{x} ,\overline{y},    z        \}$};
				\node [] (L18) at (12, -1) {$\{\overline{x}, \overline{y}              \}$};
				\node [] (L19) at (8, -3) {$\{\overline{x}                     \}$};
				\node [] (L20) at (-1, -5) {$\bot$};
				\color{black}
			\end{pgfonlayer}
			\begin{pgfonlayer}{edgelayer}
				\draw [->] (A5) to (L10);
				\draw [->] (A6) to (L10);
				\draw [->] (A1) to (L11);
				\draw [->] (L10) to (L11);
				\draw [->] (A3) to (L12);
				\draw [->] (A6) to (L12);
				\draw [->] (L12) to (L13);
				\draw [->] (A7) to (L13);
				\draw [->] (L11) to (L14);
				\draw [->] (L13) to (L14);
				\draw [->] (A3) to (L15);
				\draw [->] (A8) to (L15);
				\draw [->] (L15) to (L16);
				\draw [->] (A2) to (L16);
				\draw [->] (A5) to (L17);
				\draw [->] (A9) to (L17);
				\draw [->] (L17) to (L18);
				\draw [->] (A4) to (L18);
				\draw [->] (L16) to (L19);
				\draw [->] (L18) to (L19);
				\draw [->] (L14) to (L20);
				\draw [->] (L19) to (L20);
			\end{pgfonlayer}
		\end{tikzpicture}
		\caption{A shortest resolution refutation of the
                  formula $G$ from
			  Example~\ref{ex:decomposition-suboptimal} (axioms highlighted).}
		\label{fig:refu20}
	\end{figure}

      \end{example}

Thus concludes our tour of irreducibility and its strong cousin.
In the following subsection, we show that asymptotic resolution hardness of hitting formulas is decided on irreducible formulas---if they have polynomial-size resolution refutations, so do all unsatisfiable hitting formulas.
Then, in Subsection~\ref{subsec:irreducibility-constraint}, we employ strong irreducibility to improve the computation of shortest refutations.

\subsection{Irreducible Formulas and Asymptotic Hardness}

The goal of this subsection is to show that irreducible
hitting formulas are no easier (asymptotically up to a polynomial
factor) for resolution than non-irreducible ones, thus showing that
the question of whether hitting formulas have polynomial refutations can be
decided solely by looking at irreducible formulas.  Even though our
focus is on hitting formulas here, there is nothing special about
hittingness that makes (strongly) irreducible formulas asymptotically
hardest.  We can state and prove the theorem generally for any class
of unsatisfiable formulas that is closed under factorisation.

\begin{definition}
	Let $\cset$ be a set of unsatisfiable formulas.
	We say $\cset$ is \emph{closed under factorisation} if for every weakly reducible formula $G \in \cset$ and pseudo-factor $F$ of $G$ with clause interpolant $C$, both $F[\n{C}]$ and $G \setminus F \cup \{C\}$ are in $\cset$.
\end{definition}

\begin{theorem}
	\label{thm:polysize-irreducible}
	Let $\cset$ be a set of unsatisfiable formulas closed under factorisation, let $\sir{\cset} \subseteq \cset$ be the set of strongly irreducible formulas of $\cset$, and let $\mu : \mathbb{N} \rightarrow \mathbb{N}$ be a non-decreasing function such that every formula $F \in \sir{\cset}$ has a resolution refutation of length at most $\mu(|F|)$.
	Then every formula $F \in \cset$ has a resolution refutation of length at most $(|F|-1)\mu(|F|) + 1$.
\end{theorem}
\begin{proof}
	By induction on the number of clauses.

	\textbf{Base case.}
	Every unsatisfiable formula has at least one clause.
	The induction hypothesis holds for the only $1$-clause unsatisfiable formula, which contains only the empty clause and has a refutation of length $(1-1)\mu(1) + 1 = 1$.
	The induction hypothesis also holds for all $G \in \sir{\cset}$ because when $|G| > 1$, we have $(|G|-1)\mu(|G|) + 1 > \mu(|G|)$.

	\textbf{Inductive step.}  Consider a weakly reducible
        $G \in \cset \setminus \sir{\cset}$. Pick a non-trivial
        pseudo-factor $F \subset G$, let $C = \bigcap F$ be the clause
        interpolant of $F$ and $G \setminus F$, let $F' = F[\n{C}]$
        and $H = G \setminus F \cup \{C\}$.  By Corollary~\ref{cor:hardness-bound},
		$G$ has a refutation of size $h(F') + h(H) - 1$. Because $\cset$ is closed under
        factorisation, $F', H \in \cset$.  Both $F'$ and $H$ have
        fewer clauses than $G$, and so by the induction hypothesis
        $h(F') \leq (|F'|-1)\mu(|F'|)+1$ and
        $h(H) \leq (|H|-1)\mu(|H|) + 1$.  Summing up, we get that
        $h(G) \leq h(F') + h(H) - 1 \leq (|F'|+|H|-2)\mu(|G|) + 2 - 1
        = (|G|-1)\mu(|G|) + 1$.
\end{proof}

When $\cset = \text{UH}$, \Theorem{polysize-irreducible} translates to the following corollary.

\begin{corollary}
	\label{cor:riuh-hard}
	If irreducible unsatisfiable hitting formulas have polynomial-size resolution refutations, then so do all unsatisfiable hitting formulas.
\end{corollary}
\begin{proof}
	In order to apply \Theorem{polysize-irreducible} to hitting formulas and obtain the statement, all that needs to be shown is that the set of hitting formulas is closed under factorisation.
	This follows from Lemma~33 of \cite{KullmannZhao16} (and from the fact that any pseudo-factor of a hitting formula is a factor by \Theorem{strongly-irreducible}), but we also repeat the argument below for sake of completeness.

	It is easy to see that if $G$ is hitting, then $F \subseteq G$ is also hitting, and hittingness is further preserved under restriction, hence $F[\n{C}]$ is hitting.
	Now, consider the other formula $G \setminus F \cup \{C\}$, where $F$ is a factor of $G$ and $C$ is the basis of $F$.
	Clearly $G \setminus F \subseteq G$ is hitting.
	It remains to show that $C$ clashes with all clauses in $G \setminus F$.
	By contradiction, assume it does not clash with $D \in G \setminus F$.
	Then, because $G$ is hitting, $D$ clashes with every clause in $F$, but because $D$ does not clash with $C$, in fact $D$ clashes with every clause in $F[\n{C}] = \{E \setminus C \SM E \in F\}$.
	This is the same as saying that $\n{D}$ is a satisfying assignment for $F[\n{C}]$, contradicting the assumption that $F \models C$. 
\end{proof}

In fact, a slightly stronger statement still than Corollary~\ref{cor:riuh-hard} can be derived from \Theorem{polysize-irreducible}.
Let $(G_1, \dots)$ be an infinite sequence of hitting formulas of super-polynomial resolution hardness, assuming one exists.
Let $\cset$ be the closure of $\{G_1, \dots\}$ under factorisation, i.e., the intersection of all sets closed under factorisation containing $\{G_1, \dots\}$.
Then $\cset$ contains an infinite sequence of irreducible formulas of superpolynomial hardness---for if it did not, then \Theorem{polysize-irreducible} would imply that the sequence $(G_1, \dots)$ also has polynomial-size refutations.
In other words, in order to construct a family of hard hitting formulas, one must at least implicitly construct a family of hard irreducible hitting formulas.

\section{Symmetries}

In this section we will see how formula symmetries can be used to manipulate resolution refutations and to single out certain canonical ones.

%
\begin{lemma}
	\label{lem:proof-symmetry}
	Let $G$ be an unsatisfiable formula, $\phi \in \symg{G}$ a symmetry with $\phi^*(x) = y$ for some $x, y \in \var(G)$.
	If there is a shortest refutation of $G$ that ends with resolving $x$, then there is one that ends with resolving $y$.
Moreover, the refutation can be arranged such that it ends with either the sequence $\{y\}, \{\n{y}\}, \bot$ or $\{\n{y}\}, \{y\}, \bot$ (both are always possible).
\end{lemma}
\begin{proof}
	The first part follows by applying $\sigma$ to the refutation (literal by literal).
	For the rest we need to show that neither of $\{y\}, \{\n{y}\}$ is needed to derive the other.
	Suppose $\{y\}$ is used in the derivation of $\{\n{y}\}$; surely by removing the use of $\{y\}$ we will end up with a derivation of a clause that at most contains an additional $\n{y}$---but it already did to begin with.
\end{proof}

\begin{lemma}
	\label{lem:representative-refutation}
	Let $G$ be an unsatisfiable formula, $\symg{G}$ its symmetry group, $\orb$ the set of orbits under the variable action.
	Let $R \subseteq \var(G)$ be a set of representatives of each orbit from $\orb$, i.e., for all $O \in \mathcal{O} \; |R \cap O| = 1$.
	Then there is a variable $v \in R$ and a shortest refutation of $G$
        which ends in the sequence $\{v\}, \{\n{v}\}, \bot$, and one
        which ends in the sequence $\{\n{v}\}, \{v\}, \bot$.
\end{lemma}
\begin{proof}
	There is some shortest refutation of $G$ and it ends by resolving some variable $x$.
	Let $y \in R \cap \orb(x)$, and apply \Lemma{proof-symmetry}.
\end{proof}

\lv{
\begin{corollary}
  If $G$ is variable-transitive, then for every $v \in \var(G)$ there
  is a shortest refutation of $G$ which ends in the sequence
  $\{v\}, \{\n{v}\}, \bot$, and one which ends in the sequence
  $\{v\}, \{\n{v}\}, \bot$.
\end{corollary}
\begin{proof}
	That $G$ is variable-transitive just means that every $\{v\}$ is a set of representatives.
\end{proof}
}

\section{Generating Hitting Formulas}
\label{sec:generate}

We generate hitting formulas with a tailor-made adaptation of Nauty~\cite{Nauty}, which generates clause-literal graphs modulo isomorphisms.
In order to generate formulas efficiently, we hook into Nauty's generation process and prune partially constructed formulas on the fly when we determine they cannot be extended to full formulas.

Suppose we are generating formulas with $n$ variables and $m$ clauses.
Nauty generates the formulas by progressively adding vertices corresponding to clauses in all possible ways modulo isomorphisms.
Thus, it effectively exploring a tree of `partial' formulas, and the leaf nodes are `full' formulas we want to generate.
Suppose we have a partially constructed candidate formula $F = \{C_1, \dots C_{m'}\}, m' \leq m$ (including the possibility $m'=m$, i.e., the formula is a leaf node).
Thanks to the way Nauty generates graphs, we can assume that $C_{m'}$ is the largest clause of $F$ and that all further clauses added in any extension of $F$ will be at least as large as $C_{m'}$.
As we are interested in generating only IUHs and not arbitrary formulas, we can afford to prune as follows (here \emph{prune} means discard the current partial formula and immediately backtrack in the tree explored by Nauty):

\begin{itemize}
	\item
		if $\modelcount{F} > (m-m')2^{n-|C_{m'}|}$, prune (otherwise future clauses cannot cover all satisfying assignments).
	\item
		If $\modelcount{F} < m-m'$, prune (every clause must uniquely cover at least one assignment, and here we have too few satisfying assignments to be uniquely covered by future clauses).
	\item
		If $F$ is not hitting, prune.
	\item
		If $F$ has a non-trivial factor, or if $F$ itself is a factor and $1 < m' < m$, prune.
\end{itemize}

The first three checks can be implemented in linear time: models can be counted easily thanks to \Theorem{hitting-models}, for hittingness it is sufficient to check whether the last clause hits every other, because the same check has been performed on $\{C_1, \dots, C_{m'-1}\}$ already.

We test the fourth condition by enumerating all non-empty subsets $F' \subseteq F$ and checking whether $F' \equiv I := \{\bigcap F'\}$.
Since obviously $I \models F'$, it is sufficient to compare the \emph{number} of models in order to see whether $I$ and $F'$ are equivalent---and this can be done in polynomial time as before, because $F'$ is still hitting.
Moreover, thanks to the incremental nature of the check, we can focus only on subsets $F'$ with $C_{m'} \in F'$.

We note that it is possible to test irreducibility faster than by traversing all subsets of clauses.
The key observation is that any factor in a minimally unsatisfiable formula must be \emph{intersection-maximal}, meaning that any its strict superset has a strictly smaller intersection (we leave this as exercise for the reader, it follows straightforwardly from the definitions).
Thus, one only needs to enumerate intersection-maximal subsets of clauses, which can be done with polynomial delay (enumerate subsets recursively and after each branching, include all clauses that contain the current intersection; again we leave the details to the reader).
However, we did not implement this in Nauty, as the full traversal has an advantage in technical simplicity that outweighs algorithmic gains for small numbers of clauses.

We also generated some reducible regular unsatisfiable hitting formulas (RUHs).
In that case, we can re-use the pruning described above, and simply skip the fourth test.
However, focusing on irreducible formulas allows us to prune significantly more and generate formulas faster and up to bigger size; and the resulting number of formulas is smaller making subsequent processing more manageable~(see Section~\ref{sec:results}).

\section{Computing Shortest Refutations}
\label{sec:encoding}

In this section we will see how to use (strong) irreducibility and formula symmetries in order to rule out certain refutations, thereby pruning the search space and making it easier for the solver to show no refutations of a given length exist.

The problem we are solving asks, given a formula $F$ and integer $s$,
whether $F$ has a resolution refutation of length at most $s$, i.e.,
whether there is a sequence $C_1, \ldots, C_s$ such that each $C_i$ is
either a clause of $F$, or is derived by resolution from two previous
clauses.  We will solve the problem by reduction to SAT, building on
a previous encoding of ours~\cite{PeitlSzeider21}; for a full
presentation of the encoding we refer to that paper.  Here we will
only recall that the encoding is centered around the following three
sets of propositional variables:

\begin{itemize}
	\item
		\propvar{pos}{i, v} and \propvar{neg}{i, v} denote that the $v$ and $\n{v}$ occur in the $i$-th clause of the refutation;
	\item
		\propvar{arc}{i, j} denotes that the $i$-th clause is used to obtain the $j$-th clause via resolution.
		Every non-axiom clause has exactly two incoming arcs, i.e., for all $j > m \; \sum_{i=1}^{j-1} \propvar{arc}{i, j} = 2$ (where $m$ is the number of clauses).
\end{itemize}

An assignment to these variables together fully determines a candidate resolution derivation: it is then left to constraints to validate that the derivation is a valid refutation.

We generally assume that $F$ is minimally unsatisfiable, as all formulas we work with are.
In the following two subsections, we first show how we can add a further redundant constraint when we additionally assume strong irreducibility, and then how we can reason about refutation symmetries via formula symmetries.

\subsection{Strong Irreducibility and Clause Reuse}
\label{subsec:irreducibility-constraint}

Recall \Lemma{axiom-reuse}, which tells us that we must resolve some axioms more than once.
We will simply translate it into a SAT encoding.
We define a set of fresh variables \propvar{active}{i, j} equal to the clause $\{\propvar{arc}{i, j}, \dots, \propvar{arc}{i, s}\}$
to denote that the $i$-th clause is still `active' at position $j$ in the refutation, i.e., it is used for resolution to obtain the $j$-th or later clause.%
\footnote{We use the Tseitin encoding in order to translate this set of equations into CNF.}
With this, we can rewrite \Lemma{axiom-reuse} right into the set of clauses
\[ \big\{\n{\propvar{arc}{i, k}}, \n{\propvar{arc}{j, k}}, \propvar{active}{i, k+1}, \propvar{active}{j, k+1} \big\}, \]
for $1 \leq i < j \leq m$ and $m < k < s$. 
In practice, because this constraint can get quite large, we decided to use only a limited version, where $k$ is fixed to $m+1$ (i.e., we only apply \Lemma{axiom-reuse} to the first resolvent of the refutation).

We point out that while the constraint just described is a translation of \Lemma{axiom-reuse}, successful application also relies on \Theorem{strongly-irreducible}.
That is because while we require strong irreducibility, we generate formulas that are only irreducible.
But because they are hitting (saturated minimally unsatisfiable), \Theorem{strongly-irreducible} kicks in and warrants strong irreducibility.

\subsection{Symmetries}

Recall \Lemma{representative-refutation}, which says that formula symmetries can be applied in order to obtain symmetric images of refutations.
If we fix an arbitrary set $R$ of variable representatives, thanks to \Lemma{representative-refutation} we need only look for refutations which end in a variable from $R$.
Further, we can place the two clashing unit clauses at the end of the refutation in the order we pick (say the penultimate clause has a negative literal). 
This is quite straightforward to put into constraints:
for all $v \in \var(F)$ we include the unit clauses $\{\n{\propvar{pos}{s-1, v}}\}$ and $\{\n{\propvar{neg}{s-2, v}}\}$, and for all $v \in \var(F) \setminus R$, we include the unit clauses $\{\n{\propvar{neg}{s-1, v}}\}$ and $\{\n{\propvar{pos}{s-2, v}}\}$.
Additionally, we stipulate that the literals appearing in $C_{s-1}$ and $C_{s-2}$ are on the same variable, i.e., for all $v \in R$, $\propvar{neg}{s-1, v} = \propvar{pos}{s-2, v}$, and the arc structure at the end of the refutation: $\{\propvar{arc}{s-2, s}\}$, $\{\propvar{arc}{s-1, s}\}$, and $\{\n{\propvar{arc}{s-2, s-1}}\}$.

The encoding presented in~our previous work~\cite{PeitlSzeider21} uses
symmetry breaking to allow only one sequence for any given refutation
DAG---the canonical topological sort.  Care must be taken to ensure
that our new formula symmetry breaking is compatible with canonical
topological sorting.  Fortunately, there is a simple way out: disable
canonical topological sorting on the clauses $C_{s-1}, C_{s-2}$.

\section{Results}
\label{sec:results}

\newcommand{\ths}{\hspace{0.7em}}

We will now present our experimental results and connections to the work of others.

The main results are summarised in Tables~\ref{table:IUH-hardness} and \ref{table:RUH-hardness}.
For each pair $n$, $m$, they show the maximum hardness of IUHs (Table~\ref{table:IUH-hardness}) and RUHs (Table~\ref{table:RUH-hardness}) with
$n$ variables and $m$ clauses, together with the number of formulas
which attain maximum hardness (subscript) and total number of formulas
with $n$ variables and $m$ clauses (superscript; both numbers modulo
isomorphisms).
We computed these numbers (the shortest refutations mainly) on a cluster of heterogeneous machines with the SAT solver CaDiCaL\footnote{\url{http://fmv.jku.at/cadical}}, which we already observed earlier to be most effective for finding shortest refutations~\cite{PeitlSzeider21}.

\begin{figure}
	\includegraphics[width=\textwidth]{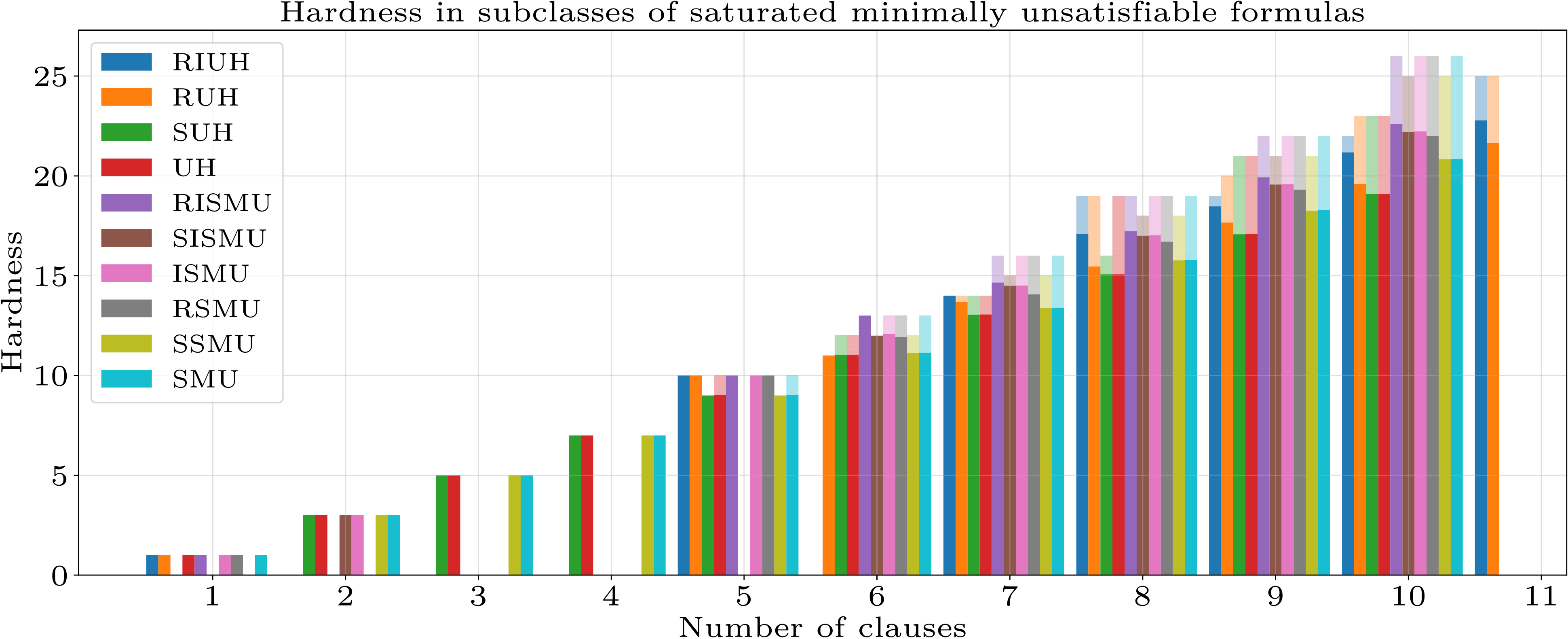}
	\caption{
		Hardness in subclasses of SMU.
		The opaque bar shows average hardness for a given number of clauses, the transparent bar extends to maximum hardness.
		While irreducible hitting formulas are not as hard as general hitting formulas in the worst case, they are harder on average.
}
		\label{fig:hardness-classes}
\end{figure}

Figure~\ref{fig:hardness-classes} compares IUH hardness to hardness of RUHs and other related classes like SMU, depicting both maximum as well as average hardness.%
\footnote{We calculate average hardness as weighted average by the number of isomorphic copies of each formula.
\lv{	As a side note for readers familiar with group theory: the number of different isomorphic copies can be computed by the orbit-stabilizer lemma applied to the case of the orbit under the group of all negation-preserving permutations of literals, with the stabilizer being the automorphism group.}
}
The data on SMU hardness is taken from our previous work~\cite{PeitlSzeider21}, where we looked for \emph{resolution hardness numbers} (maximum resolution complexity with a fixed number of clauses) and showed that the hardest formulas with a given number of clauses are saturated minimally unsatisfiable, and consequently analyzed hardness in SMU formulas with up to $10$ clauses in detail.

We can draw several conclusions.

Firstly, it is not true that IUHs are always the hardest of hitting formulas for a given number of clauses.
On the other hand, IUHs exhibit greater hardness \emph{on average} than all hitting formulas, suggesting that irreducibility is indeed positively correlated with hardness.

Secondly, with the caveat that our limited study cannot provide answers for larger formulas,
is seems that hitting formulas, irreducible or
not, are not substantially easier than formulas in general.
Interestingly, while for most values of $m$ (number of clauses) the hardest formulas are not hitting, for $5$ and $8$ clauses there is a IUH that is as hard as any formula (for $m=5$ it is $\MUtwo{5}$, for $m=8$ it is the formula $F$ from Example~\ref{ex:decomposition-suboptimal}).

Of course,
limited conclusions should be drawn from our limited computational
evaluation.  We note that it is probably not possible to advance to
much higher numbers of clauses with this approach: we are looking at
over 200K IUHs with 15 clauses, a number that is likely going to be
very difficult to process for shortest refutations.  The situation is even
worse for RUHs, with almost 140K with $12$ clauses and over 1M formulas with $13$ clauses.
A further complication is that the computation of shortest refutations grows hard quickly with increasing formula size~\cite{PeitlSzeider21}.
It already took almost $9$ days to get the hardness of the hardest IUHs with $14$ clauses whose hardness we could compute, and that still leaves many formulas with $14$ clauses left.
As we noted in our previous work~\cite{PeitlSzeider21}, this workflow is difficult to parallelize as most of the effort is concentrated in one hardest SAT query.

\begin{table}
	\begin{center}
\begin{tabular}{
		|@{\ths}l@{\ths}|@{\ths}l@{\ths}l@{\ths}l@{\ths}l@{\ths}c@{\ths}c@{\ths}l@{\ths}l@{\ths}l@{\ths}l@{\ths}l@{\ths}l@{\ths}l@{\ths}l@{\ths}|
}
\hline
  $n$\textbackslash $m$
   &    1    & 2 & 3 & 4 &     5    &  6  &     7    &    8     &    9        &     10      &     11          &     12          &     13           & 14            \\ 
\hline

0  & $1^1_1$ &   &   &   &          &     &          &          &             &             &                 &                 &                  &               \\ 
                                                                                                                                                  
1  &         &   &   &   &          &     &          &          &             &             &                 &                 &                  &               \\ 
                                                                                                                                                       
2  &         &   &   &   &          &     &          &          &             &             &                 &                 &                  &               \\ 
                                                                                                                                                       
3  &         &   &   &   & $10^1_1$ & \phantom{$10^1_1$}&  &    &             &             &                 &                 &                  &               \\ 
                                                                                                                                                       
4  &         &   &   &   &          &     & $14^2_2$ & $19_1^2$ & $19_1^1$    &             &                 &                 &                  &               \\ 
                                                                                                                                       
5  &         &   &   &   &          &     &          &          & $19_7^{15}$ & $22_9^{47}$ & $25_{15}^{138}$ & $28_2^{245}$    & $30_{11}^{511}$  & ${\scriptscriptstyle \geq} 33_?^{804}$   \\ 
                                                                                                                                    
6  &         &   &   &   &          &     &          &          &             &             & $24_6^{112}$    & $27_{10}^{618}$ & $30_{34}^{5178}$ & ${\scriptscriptstyle \geq} 34_?^{25235}$ \\ 
                                                                                                                                          
7  &         &   &   &   &          &     &          &          &             &             &                 &                 & $29_{1}^{1019}$  & ${\scriptscriptstyle \geq} 31_?^{7765}$  \\ 
                                                                                                                                                       
8  &         &   &   &   &          &     &          &          &             &             &                 &                 &                  &               \\ 
                                                                                                                                                       
9  &         &   &   &   &          &     &          &          &             &             &                 &                 &                  &               \\ 

10 &         &   &   &   &          &     &          &          &             &             &                 &                 &                  &               \\ 

          
	\hline
	\end{tabular}
	\end{center}
	\caption{Values of $h(\text{IUH}(n,m))$, i.e., the lengths of the
		longest shortest refutation required by a (regular) irreducible unsatisfiable hitting formula with $n$ variables and $m$ clauses.
		The subscript gives the number of formulas that attain maximum hardness, the superscript is the total number of formulas in each category.
		  All counts are modulo isomorphisms.
		  Empty areas contain no IUHs.
	  }
	\label{table:IUH-hardness}
\end{table}

\begin{table}
	\begin{center}
\begin{tabular}{
		|@{\ths}l@{\ths}|@{\ths}l@{\ths}l@{\ths}l@{\ths}l@{\ths}l@{\ths}l@{\ths}l@{\ths}l@{\ths}l@{\ths}l@{\ths}l@{\ths}|
}
\hline
  $n$\textbackslash $m$
   &    1    & 2 & 3 & 4       &     5    &  6       &     7       &    8        &    9         &     10        &     11           \\
\hline

0  & $1^1_1$ &   &   &         &          &          &             &             &              &               &                  \\
                                                                                                                                   
1  &         &   &   &         &          &          &             &             &              &               &                  \\
                                                                                                                             
2  &         &   &   & $7_1^1$ &          &          &             &             &              &               &                  \\
                                                                                                                        
3  &         &   &   &         & $10^1_1$ & $11^3_3$ & $13_1^1$    & $15_1^1$    &              &               &                  \\
                                                                                                                       
4  &         &   &   &         &          &          & $14^6_{10}$ & $19_1^{49}$ & $20_1^{79}$  & $21_5^{94}$   & $22_{21}^{70}$   \\
                                                                                                                         
5  &         &   &   &         &          &          &             & $16_5^9$    & $19_7^{207}$ & $23_3^{1772}$ & $25_{21}^{8203}$ \\
                                                                                                                               
6  &         &   &   &         &          &          &             &             & $19_2^4$     & $21_7^{281}$  & $25_3^{7449}$    \\
                                                                                                                                   
7  &         &   &   &         &          &          &             &             &              & $21_1^1$      & $23_{50}^{261}$  \\
                                                                                                                                   
8  &         &   &   &         &          &          &             &             &              &               &                  \\
                                                                                                                                   
9  &         &   &   &         &          &          &             &             &              &               &                  \\
	\hline
	\end{tabular}
	\end{center}
	\caption{Values of $h(\text{RUH}(n,m))$, i.e., the lengths of the
		longest shortest refutation required by a regular unsatisfiable hitting formula with $n$ variables and $m$ clauses.
		The subscript gives the number of formulas that attain maximum hardness, the superscript is the total number of formulas in each category.
		  All formula counts are modulo isomorphisms.
		  Empty areas contain no RUHs.
	  }
	\label{table:RUH-hardness}
\end{table}

On the flip side though, the pace of growth suggests there might indeed be infinitely many IUHs.
A refined existence question could ask whether IUHs with each number of clauses exist.
Table~\ref{table:IUH-hardness} shows that is not the case: IUHs with $2$, $3$, $4$, and $6$ clauses are missing.
For $2$ and $3$, regularity is the culprit: there are no regular minimally unsatisfiable formulas with $2$ or $3$ clauses either~\cite{KullmannZhao13,PeitlSzeider21}.
The case of $4$ and $6$ is a different story: there are both RUHs and RISMUs with $6$ clauses, and there is a RUH with $4$ clauses, but interestingly no IUH with $6$ clauses, and not even a RISMU with $4$ clauses.
This reminds us of other algebraic objects like orthogonal Latin squares, which also exist for almost all sizes (cf.\ OEIS sequences \href{http://oeis.org/A160368}{A160368}, \href{http://oeis.org/A305570}{A305570}, \href{http://oeis.org/A305571}{A305571}, and \href{http://oeis.org/A287761}{A287761}), but it is unclear whether there is any connection.

Kullmann and Zhao~\cite{KullmannZhao16} were interested in the existence of RUHs, and in particular they investigated the `Finiteness Conjecture' that for any fixed deficiency only finitely many RUHs exist.
An exact, stronger variant of their conjecture~\cite[Conjecture 5]{KullmannZhao16} states explicitly that a RUH of deficiency $k \geq 2$ can have at most $4k-5$ variables (and thus $5k-5$ clauses), which they prove for $k=3$ (and which was already known for $k=2$~\cite{KleineBuning00}).
This exactness allows our work to intersect with theirs, and indeed, wherever we touch the bound of $4k-5$, Tables~\ref{table:IUH-hardness} and \ref{table:RUH-hardness} confirm that no RUHs (or IUHs) of deficiency $k$ and with more than $4k-5$ variables exist.
In fact, all three extremal cases that we cover (for $k=1,2,3, n=0,3,7$) have a unique RUH formula modulo isomorphism (the unique formula for $k=3$ is the formula $K_2$ described by Kullmann and Zhao~\cite[Lemma 4]{KullmannZhao16}).
This was already well known for $k=1,2$, and followed for $k=3$ from our previous work~\cite{PeitlSzeider21}, where we generated all RSMU formulas with $7$ variables and $10$ clauses, but we did not see the connection to Kullmann and Zhao's work then.
There is another intriguing pattern of similar nature apparent in Table~\ref{table:IUH-hardness}: formulas with $n$ variables show up starting with $m=2n-1$ clauses.
We leave it to future work to find out whether these patterns are coincidental or not.

Our catalog of RUHs and IUHs will be added to our existing catalog of SMU formulas~\cite{smu}, and will hopefully help with the testing of hypotheses and as inspiration for constructing infinite classes of irreducible hitting formulas.
In Figure~\ref{fig:selected-formulas}, we show hand-picked examples of notable formulas.
For each number of clauses, we picked among the IUHs with highest hardness one with the smallest length (number of literal occurrences, i.e.\ $\sum_{C\in F} |C|$), and in case there was more than one, the one with the largest symmetry group.
\lv{These are solely heuristics meant to maximize aesthetic pleasure---by minimizing the amount displayed and maximizing its symmetry.}

\definecolor{auburn}{rgb}{0.43, 0.21, 0.1}
\definecolor{azure}{rgb}{0.0, 0.5, 1.0}
\definecolor{burgundy}{rgb}{0.5, 0.0, 0.13}
\definecolor{cadmiumred}{rgb}{0.89, 0.0, 0.13}
\definecolor{carminered}{rgb}{1.0, 0.0, 0.22}
\definecolor{ceruleanblue}{rgb}{0.16, 0.32, 0.75}
{
	\newcommand{\m}{\textcolor{carminered}{\textbf{--}}}
	\newcommand{\p}{\textcolor{ceruleanblue}{\textbf{+}}}
\def\arraystretch{0.9}
\begin{figure}[htb]
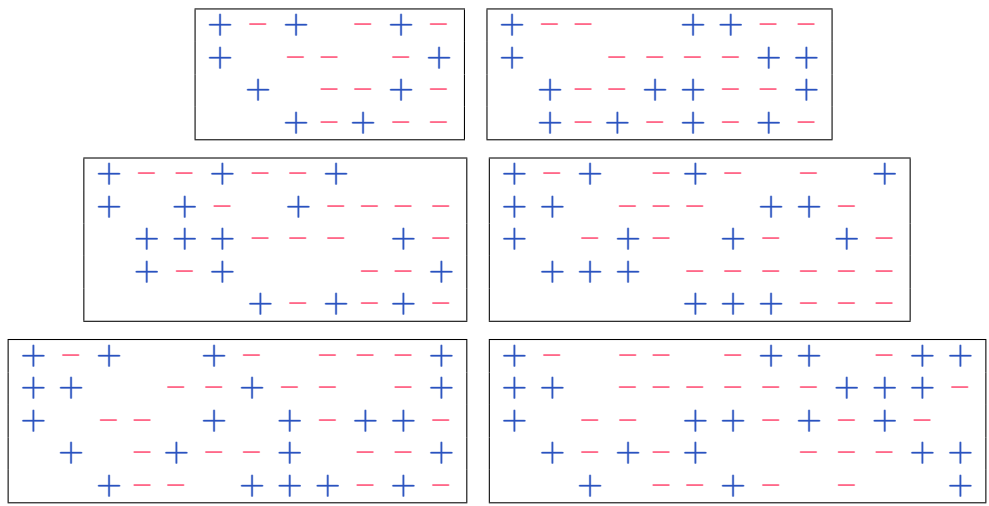

     	\centering

		\hspace{0.8em}
	\begin{tabular}{|@{\sep}c@{\sep}c@{\sep}c@{\sep}c@{\sep}c@{\sep}c@{\sep}c@{\sep}|}
		\hline
		\p & \m & \p &    & \m & \p & \m \vsep
		\p &    & \m & \m &    & \m & \p \vsep
		   & \p &    & \m & \m & \p & \m \vsep
		   &    & \p & \m & \p & \m & \m \vsep
		\hline
	\end{tabular}
	\sep
	\begin{tabular}{|@{\sep}c@{\sep}c@{\sep}c@{\sep}c@{\sep}c@{\sep}c@{\sep}c@{\sep}c@{\sep}c@{\sep}|}
		\hline
		\p & \m & \m &    &    & \p & \p & \m & \m \vsep
		\p &    &    & \m & \m & \m & \m & \p & \p \vsep
		   & \p & \m & \m & \p & \p & \m & \m & \p \vsep
		   & \p & \m & \p & \m & \p & \m & \p & \m \vsep
		\hline
	\end{tabular}
	\\[0.5em]
	\begin{tabular}{|@{\sep}c@{\sep}c@{\sep}c@{\sep}c@{\sep}c@{\sep}c@{\sep}c@{\sep}c@{\sep}c@{\sep}c@{\sep}|}
		\hline
		\p & \m & \m & \p & \m & \m & \p &    &    &    \vsep
		\p &    & \p & \m &    & \p & \m & \m & \m & \m \vsep
		   & \p & \p & \p & \m & \m & \m &    & \p & \m \vsep
		   & \p & \m & \p &    &    &    & \m & \m & \p \vsep
		   &    &    &    & \p & \m & \p & \m & \p & \m \vsep
		\hline
	\end{tabular}
	\sep
	\begin{tabular}{|@{\sep}c@{\sep}c@{\sep}c@{\sep}c@{\sep}c@{\sep}c@{\sep}c@{\sep}c@{\sep}c@{\sep}c@{\sep}c@{\sep}|}
		\hline
		\p & \m & \p &    & \m & \p & \m &    & \m &    & \p \vsep
		\p & \p &    & \m & \m & \m &    & \p & \p & \m &    \vsep
		\p &    & \m & \p & \m &    & \p & \m &    & \p & \m \vsep
		   & \p & \p & \p &    & \m & \m & \m & \m & \m & \m \vsep
		   &    &    &    &    & \p & \p & \p & \m & \m & \m \vsep
		\hline
	\end{tabular}
	\\[0.5em]
	\begin{tabular}{|@{\sep}c@{\sep}c@{\sep}c@{\sep}c@{\sep}c@{\sep}c@{\sep}c@{\sep}c@{\sep}c@{\sep}c@{\sep}c@{\sep}c@{\sep}|}
		\hline
		\p & \m & \p &    &    & \p & \m &    & \m & \m & \m & \p \vsep
		\p & \p &    &    & \m & \m & \p & \m & \m &    & \m & \p \vsep
		\p &    & \m & \m &    & \p &    & \p & \m & \p & \p & \m \vsep
		   & \p &    & \m & \p & \m & \m & \p &    & \m & \m & \p \vsep
		   &    & \p & \m & \m &    & \p & \p & \p & \m & \p & \m \vsep
		\hline
	\end{tabular}
	\sep
	\begin{tabular}{|@{\sep}c@{\sep}c@{\sep}c@{\sep}c@{\sep}c@{\sep}c@{\sep}c@{\sep}c@{\sep}c@{\sep}c@{\sep}c@{\sep}c@{\sep}c@{\sep}|}
		\hline
		\p & \m &    & \m & \m &    & \m & \p & \p &    & \m & \p & \p \vsep
		\p & \p &    & \m & \m & \m & \m & \m & \m & \p & \p & \p & \m \vsep
		\p &    & \m & \m &    & \p & \p & \m & \p & \m & \p & \m &    \vsep
		   & \p & \m & \p & \m & \p &    &    & \m & \m & \m & \p & \p \vsep
		   &    & \p &    & \m & \m & \p & \m &    & \m &    &    & \p \vsep
		\hline
	\end{tabular}
     
	\caption{Selected hardest IUHs with $7$ and $9-13$ clauses
\sv{written in `incidence matrix' form (rows are variables, columns are clauses, `\p'  and `\m' mean positive and negative occurrence)}.
	The hardest IUH with $1$ clause is $\{\bot\}$, with $5$ clauses $\MUtwo{5}$, already shown in Example~\ref{ex:read-twice}, and with $8$ clauses the formula $F$ from Example~\ref{ex:decomposition-suboptimal}.
\lv{The matrix notation is the same as in Example~\ref{ex:decomposition-suboptimal}.}}
	\label{fig:selected-formulas}
\end{figure}
}


\section{Conclusion}

Inspired by the observation that hitting formulas are remarkable among polynomial-time decidable classes of propositional formulas in that they admit polynomial-time model counting and at the same time lack obvious resolution-complexity upper bounds, we set out to answer whether hitting formulas are hard for resolution.
This quest led us into the land of (strong) irreducibility and to discover a number of interesting related questions and phenomena.
With our theoretical and experimental results we now understand that IUHs are a key subset of hitting formulas with respect to both resolution complexity, as well as just plain existence.
Because the number of IUHs (and even more so RUHs) grows so fast, it seems safe to conjecture that there are infinitely many---but at the same time there are intriguing similarities with other algebraic objects like orthogonal Latin squares, and the rules governing existence of IUHs (and RUHs) with fixed parameters $n$ and $m$ also exhibit simple-looking patterns that deserve further investigation.
Based on the resolution complexity we could observe, we see no signs that hitting formulas should be significantly easier for resolution than formulas in general, even if they are a bit easier than the absolutely hardest formulas; although we cannot draw definitive conclusions beyond the formulas whose proofs we computed.

While we focused on resolution here, it would also be interesting to look at hitting formulas' complexity in other proof systems.

\bibliography{literature}

\begin{thebibliography}{10}

\bibitem{AharoniLinial86}
Ron Aharoni and Nathan Linial.
\newblock Minimal non-two-colorable hypergraphs and minimal unsatisfiable
  formulas.
\newblock {\em J. Combin. Theory Ser. A}, 43:196--204, 1986.

\bibitem{AtseriasFichteThurley11}
Albert Atserias, Johannes~Klaus Fichte, and Marc Thurley.
\newblock Clause-learning algorithms with many restarts and bounded-width
  resolution.
\newblock {\em J. Artif. Intell. Res.}, 40:353--373, 2011.

\bibitem{BeameKautzSabharwal04}
Paul Beame, Henry~A. Kautz, and Ashish Sabharwal.
\newblock Towards understanding and harnessing the potential of clause
  learning.
\newblock {\em J. Artif. Intell. Res.}, 22:319--351, 2004.

\bibitem{BondyMurty08}
J.~A. Bondy and U.~S.~R. Murty.
\newblock {\em Graph theory}, volume 244 of {\em Graduate Texts in
  Mathematics}.
\newblock Springer Verlag, New York, 2008.

\bibitem{DavisPutnam60}
M.~Davis and H.~Putnam.
\newblock A computing procedure for quantification theory.
\newblock {\em J. of the ACM}, 7(3):201--215, 1960.

\bibitem{FrancoMartin21}
John Franco and John Martin.
\newblock A history of satisfiabilty.
\newblock In Armin Biere, Marijn Heule, Hans van Maaren, and Toby Walsh,
  editors, {\em Handbook of Satisfiability, 2nd Ed.}, chapter~1, pages 3--74.
  IOS Press, 2021.

\bibitem{GalesiKullmann04}
Nicola Galesi and Oliver Kullmann.
\newblock Polynomial time {SAT} decision, hypergraph transversals and the
  {Hermitian} rank.
\newblock In {\em {SAT} 2004 - The Seventh International Conference on Theory
  and Applications of Satisfiability Testing, 10-13 May 2004, Vancouver, BC,
  Canada, Online Proceedings}, 2004.

\bibitem{GanianSzeider21}
Robert Ganian and Stefan Szeider.
\newblock New width parameters for {SAT} and \#{SAT}.
\newblock {\em Artificial Intelligence}, 295:103460, 2021.

\bibitem{Haken85}
Armin Haken.
\newblock The intractability of resolution.
\newblock {\em Theoretical Computer Science}, 39:297--308, 1985.

\bibitem{Iwama89}
Kazuo Iwama.
\newblock C{NF}-satisfiability test by counting and polynomial average time.
\newblock {\em SIAM J. Comput.}, 18(2):385--391, 1989.

\bibitem{KleineBuning00}
Hans {Kleine B\"uning}.
\newblock On subclasses of minimal unsatisfiable formulas.
\newblock {\em Discr. Appl. Math.}, 107(1--3):83--98, 2000.

\bibitem{KleineBuningZhao01a}
Hans {Kleine B\"uning} and Xishun Zhao.
\newblock Satisfiable formulas closed under replacement.
\newblock In Henry Kautz and Bart Selman, editors, {\em Proceedings for the
  Workshop on Theory and Applications of Satisfiability}, volume~9 of {\em
  Electronic Notes in Discrete Mathematics}. Elsevier Science Publishers,
  North-Holland, 2001.

\bibitem{KleineBuningZhao02b}
Hans Kleine{ }B{\"{u}}ning and Xishun Zhao.
\newblock The complexity of read-once resolution.
\newblock {\em Ann. Math. Artif. Intell.}, 36(4):419--435, 2002.

\bibitem{KleineBuningZhao03}
Hans Kleine~B{\"u}ning and Xishun Zhao.
\newblock On the structure of some classes of minimal unsatisfiable formulas.
\newblock {\em Discr. Appl. Math.}, 130(2):185--207, 2003.

\bibitem{Kullmann03}
Oliver Kullmann.
\newblock The combinatorics of conflicts between clauses.
\newblock In Enrico Giunchiglia and Armando Tacchella, editors, {\em Sixth
  International Conference on Theory and Applications of Satisfiability
  Testing, S. Margherita Ligure - Portofino, Italy, May 5-8, 2003 (SAT 2003)},
  volume 2919 of {\em Lecture Notes in Computer Science}. Springer Verlag,
  2004.

\bibitem{Kullmann10}
Oliver Kullmann.
\newblock Green-{Tao} numbers and {SAT}.
\newblock In Ofer Strichman and Stefan Szeider, editors, {\em Theory and
  Applications of Satisfiability Testing - {SAT} 2010, 13th International
  Conference, {SAT} 2010, Edinburgh, UK, July 11-14, 2010. Proceedings}, volume
  6175 of {\em Lecture Notes in Computer Science}, pages 352--362. Springer,
  2010.

\bibitem{Kullmann11}
Oliver Kullmann.
\newblock Constraint satisfaction problems in clausal form {II}: {M}inimal
  unsatisfiability and conflict structure.
\newblock {\em Fund. Inform.}, 109(1):83--119, 2011.

\bibitem{KullmannZhao13}
Oliver Kullmann and Xishun Zhao.
\newblock On {D}avis-{P}utnam reductions for minimally unsatisfiable
  clause-sets.
\newblock {\em Theoretical Computer Science}, 492:70--87, 2013.

\bibitem{KullmannZhao16}
Oliver Kullmann and Xishun Zhao.
\newblock Unsatisfiable hitting clause-sets with three more clauses than
  variables.
\newblock {\em CoRR}, abs/1604.01288, 2016.

\bibitem{Nauty}
Brendan~D. McKay and Adolfo Piperno.
\newblock Practical graph isomorphism, \{II\}.
\newblock {\em Journal of Symbolic Computation}, 60(0):94 -- 112, 2014.

\bibitem{NishimuraRagdeSzeider07}
Naomi Nishimura, Prabhakar Ragde, and Stefan Szeider.
\newblock Solving \#{S}{A}{T} using vertex covers.
\newblock {\em Acta Informatica}, 44(7-8):509--523, 2007.

\bibitem{OrdyniakPaulusmaSzeider13}
Sebastian Ordyniak, Dani{\"e}l Paulusma, and Stefan Szeider.
\newblock Satisfiability of acyclic and almost acyclic {CNF} formulas.
\newblock {\em Theoretical Computer Science}, 481:85--99, 2013.

\bibitem{PeitlSzeider21}
Tom{\'{a}}{\v{s}} Peitl and Stefan Szeider.
\newblock Finding the hardest formulas for resolution.
\newblock {\em J. Artif. Intell. Res.}, 72:69--97, 2021.
\newblock Conference Award Track, best paper CP 2020.

\bibitem{smu}
Tomáš Peitl and Stefan Szeider.
\newblock {\em Saturated Minimally Unsatisfiable Formulas on up to Ten
  Clauses}.
\newblock Zenodo, January 2021.
\newblock https://doi.org/10.5281/zenodo.3951545.

\bibitem{PipatsrisawatDarwiche09}
Knot Pipatsrisawat and Adnan Darwiche.
\newblock On the power of clause-learning {SAT} solvers with restarts.
\newblock In Ian~P. Gent, editor, {\em Principles and Practice of Constraint
  Programming - {CP} 2009, 15th International Conference, {CP} 2009, Lisbon,
  Portugal, September 20-24, 2009, Proceedings}, volume 5732 of {\em Lecture
  Notes in Computer Science}, pages 654--668. Springer Verlag, 2009.

\bibitem{Roth96}
Dan Roth.
\newblock On the hardness of approximate reasoning.
\newblock {\em Artificial Intelligence}, 82(1-2):273--302, 1996.

\bibitem{MarquessilvaSakallah96}
Jo{\~a}o P.~Marques Silva and Karem~A. Sakallah.
\newblock {GRASP} - a new search algorithm for satisfiability.
\newblock In {\em International Conference on Computer-Aided Design (ICCAD
  '96), November 10-14, 1996, San Jose, CA, USA}, pages 220--227. ACM and IEEE,
  1996.

\bibitem{Szorenyi08}
Bal{\'{a}}zs Sz{\"{o}}r{\'{e}}nyi.
\newblock Disjoint {DNF} tautologies with conflict bound two.
\newblock {\em J on Satisfiability, Boolean Modeling and Computation},
  4(1):1--14, 2008.

\bibitem{Urquhart87}
Alasdair Urquhart.
\newblock Hard examples for resolution.
\newblock {\em J. of the ACM}, 34(1):209--219, 1987.

\bibitem{Zhao06}
Xishun Zhao.
\newblock Complexity results on minimal unsatisfiable formulas.
\newblock In {\em Mathematical logic in Asia}, pages 302--319. World Sci.
  Publ., Hackensack, NJ, 2006.

\end{thebibliography}

\end{document}